\documentclass{article}

\usepackage[numbers]{natbib}

\usepackage[final]{nips_2017}

\usepackage[utf8]{inputenc} 
\usepackage[T1]{fontenc}    
\usepackage{hyperref}       
\usepackage{url}            
\usepackage{booktabs}       
\usepackage{amsfonts}       
\usepackage{nicefrac}       
\usepackage{microtype}      
\usepackage{graphicx}

\title{Learning Neural Markers of Schizophrenia Disorder Using Recurrent Neural Networks}

\author{
Jumana Dakka\\
Electrical \& Computer Engineering\\
Rutgers University\\
\texttt{jumana.dakka@rutgers.edu} \\
\And
Pouya Bashivan\\
Brain and Cognitive Sciences\\
Massachusetts Institute of Technology\\
\texttt{bashivan@mit.edu} \\
\And
Mina Gheiratmand\\
Computing Science\\
University of Alberta\\
\texttt{mgheirat@ualberta.ca}\\
\And
Irina Rish\\
IBM T.J. Watson Research Center\\
\texttt{rish@us.ibm.com}\\
\And
Shantenu Jha\\
Electrical \& Computer Engineering\\
Rutgers University\\
\texttt{shantenu.jha@rutgers.edu}\\
\And
Russell Greiner\\
Computing Science\\
University of Alberta\\
\texttt{rgreiner@ualberta.ca}\\
}

\begin{document}

\maketitle

\begin{abstract}


Smart systems that can accurately diagnose patients with mental disorders and identify effective treatments based on brain functional imaging data are of great applicability and are gaining much attention. Most previous machine learning studies use hand-designed features, such as functional connectivity, which does not maintain the potential useful information in the spatial relationship between brain regions and the temporal profile of the signal in each region. Here we propose a new method based on recurrent-convolutional neural networks to automatically learn useful representations from segments of 4-D fMRI recordings. Our goal is to exploit both spatial and temporal information in the functional MRI \textit{movie} (at the whole-brain voxel level) for identifying patients with schizophrenia. 

\end{abstract}

\section{Introduction}

Diagnosis of psychiatric diseases is challenging as there are currently no objective biological markers associated with mental disorders. Similarity of symptoms among different diseases (e.g. depression phase of bipolar disorder and unipolar depression) can lead to inaccurate diagnosis, and to less effective intervention. Worse, there is also no objective biological marker for predicting treatment response in an individual. This oftentimes results in multiple changes in a patient's prescription, often resulting in poor adherence given the medication's side effects. Such inefficiency in the diagnosis and treatment prognosis process for psychiatric disorders has increased the global burden of disease, with mental illness ranking first, before cancer and cardiac conditions, in terms of time lost to disability (WHO 2012 report) and costs \citep{Roehrig2016}.

In recent years, machine learning techniques have shown success in identifying patients with mental or neurological disorders and in predicting treatment response using brain imaging, especially structural and/or functional MRI (magnetic resonance imaging) data \citep{Orru2012, Zarogianni2013, Koutsouleris2016, Vieira2017, Gheiratmand2017}. Almost all these studies extract features from imaging data then apply standard learning algorithms to produce classifiers, such as support vector machines (SVM) \citep{Orru2012, Wolfers2015} that can discriminate between patients and controls, or predict response to treatment.
Some typical imaging features extracted from functional MRI (fMRI) or structural MRI (sMRI) data include functional connectivity (FC) and amplitude of low-frequency fluctuations (ALFF) for fMRI, and voxel-based morphometry and gray matter thickness/volume for sMRI. Such features may be extracted voxel-wise (where every voxel is a brain tissue of size $ \sim 1-27 mm^3$) or region-wise, from predefined brain regions (e.g. thalamus, postcentral gyrus).

With deep learning techniques providing outstanding performance in various fields, including image classification, speech recognition, and video classification, among others, this approach is being explored in clinical applications, including those involving medical imaging data \citep{Shen2017, Litjens2017, Gulshan2016}. In addition to their potential to surpass the performance of other standard machine learning techniques, deep learning  methods are attractive because they can be applied directly to the data, skipping the need to extract hand-designed features, a step that is necessary in almost all other machine learning approaches. In addition to the possibility of improving prediction accuracy, deep neural networks (DNN) allow us to move away from hypothesis-driven feature selection to data-driven feature discovery.
Various deep learning methods (e.g. multi-layer perceptron, autoencoders, deep belief networks, and convolutional neural networks) have been used to analyze imaging data for various psychiatric and neurological disorders, including but not limited to Alzheimer's disease, ADHD, and Psychosis \citep[see][for a review]{Vieira2017}. Most of these studies use sMRI for predictions in neurological disorders, and much fewer studies use fMRI \citep{Plis2014, Kim2016, Suk2016, Sarraf2016}, which has been shown to be particularly relevant in predictive analysis of psychiatric disorders (such as schizophrenia) \citep{Damaraju2014, Calhoun2009}. fMRI data measures blood oxygenation level-dependent (BOLD) signal at every brain voxel by taking a scan of the whole brain every 1-3 s. This produces a \textit{movie} of the brain activity (reflected in BOLD signal\footnote{Note the relationship between BOLD signal and neural activity is still under scrutiny \citep{RN1}.}), either in response to a task (e.g. a motor, sensory, or cognitive task) or simply at rest.

\begin{figure}[t]
\begin{center}
\includegraphics[width=2.5in]{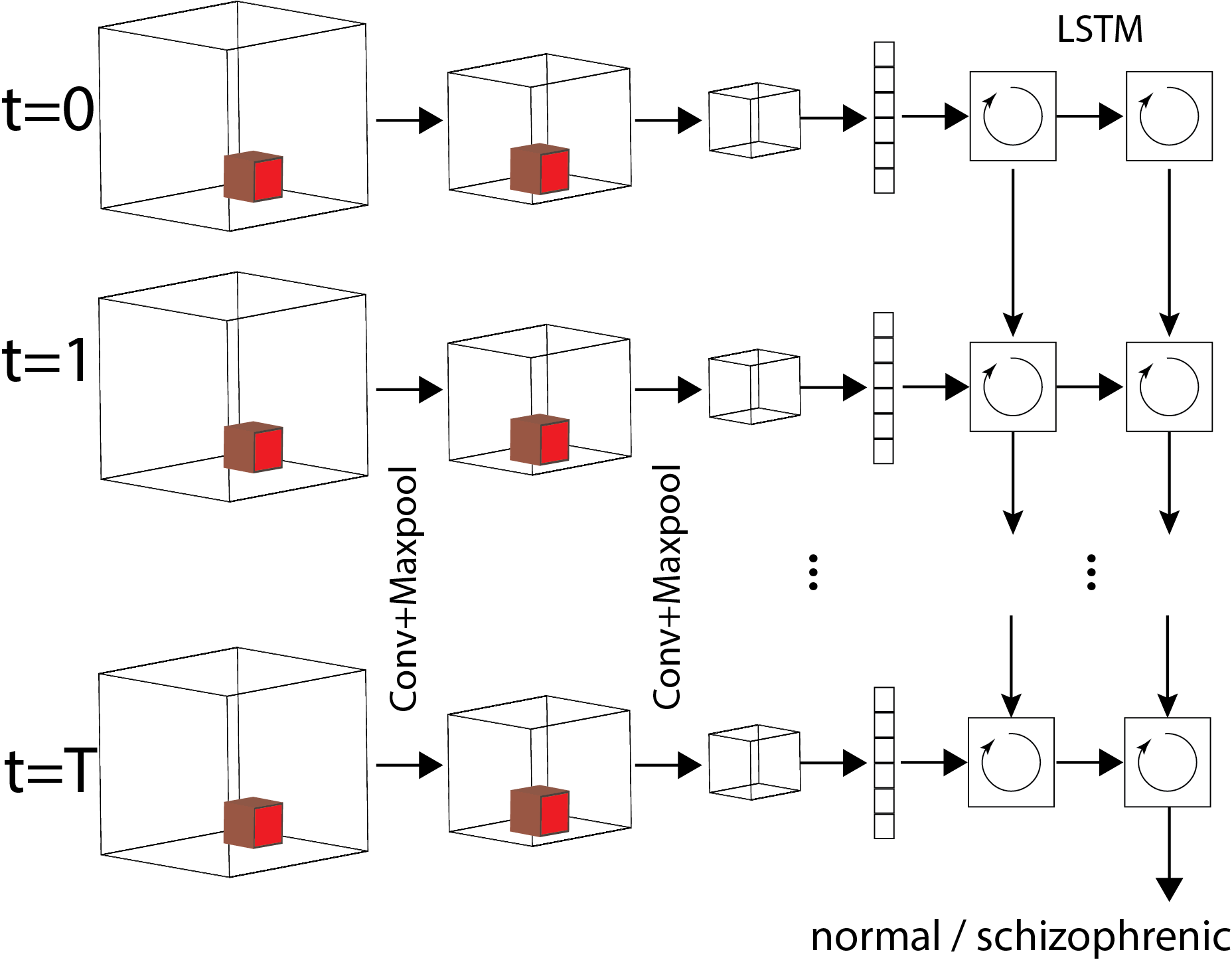}
\end{center}
\caption{Overview of the recurrent-convolutional neural network approach.}
\label{fig1}
\end{figure}

Here, our goal is to exploit both spatial and temporal information in the fMRI movie (at the whole-brain voxel level) to distinguish patients with schizophrenia vs healthy controls. We propose using a recurrent convolutional neural network (R-CNN) involving a 3-D CNN followed by a sequential neural network with LSTM (long short term memory) units. The CNN extracts spatial features, which are fed to the LSTM model, that uses the dependencies between time points at every spatial location to generate a label $\in\{patient, control\}$ (see Figure-\ref{fig1}). To our knowledge, this is the first work to apply a recurrent CNN to fMRI data for neurological/psychiatric diagnosis (here schizophrenia). As discussed earlier, most previous fMRI/machine learning studies, including some that used DNNs \citep{Kim2016}, use hand-designed features, in particular FC features \citep{Gheiratmand2017}, which collapses the time dimension into one single number (i.e., the correlation coefficient between a pair of time-series). Such approaches do not keep track of the relationships between spatial locations (e.g. voxel or brain regions) either. Here, we expand the work by \citep{Bashivan2016}, who successfully applied a R-CNN (with 2-D convolutions) to EEG data in a mental load classification task, to fMRI data (using 3-D convolutions). 
We used fMRI data in response to an auditory oddball task from patients diagnosed with schizophrenia and healthy controls from FBIRN dataset \citep{Keator2016}. The task is to predict whether this sample came from a patient, or a control, based on the preprocessed fMRI BOLD signal at the voxel level, exploiting the temporal and spatial information in the data within an end-to-end deep learning framework.

\section{Methods}

\subsection{Dataset}

We used the FBIRN phase-II fMRI dataset \citep{Keator2016}
, which includes functional and structural MRI data for patients with schizophrenia or schizoaffective disorder and age- and sex-matched healthy controls. We focused on the subset of fMRI data acquired in response to an Auditory Oddball (AO) task.
The fMRI data included whole-brain scans taken every 2 seconds for a period of 280 seconds. Per subject, there were 4 experiment runs (280 sec each).
A standard preprocessing pipeline was applied to each subject's raw fMRI data using the FSL software package \citep{Jenkinson2012}. The pipeline included motion correction, tCompCor denoising, spatial filtering, high-pass temporal filtering, and linear registration to the MNI T1 template through subject's T1 scan \citep[see][for a more detailed description of the preprocessing stages]{Gheiratmand2017}. The first 3 volumes in each run were deleted for signal instability, resulting in a total of 137 volumes. Finally, a universal mask, i.e. the intersection of all subjects' brains, was applied to each subject's data, resulting in a common non-zero brain area of 26,949 voxels (brain tissue of size $3.4374\times3.4375\times5$ mm).
After preprocessing and quality control, N = 95 subjects (46 patients, 49 controls) remained in the study (from a total of 164 subjects in both scanning sessions).(Subjects with missing imaging data, runs less than 4, or excessive motion were excluded \citep[see][for details of quality control]{Gheiratmand2017}.)

\subsection{Models}

We compared the accuracy of recurrent neural networks (LSTM model) and recurrent-convolutional neural networks (R-CNN), versus a linear and a nonlinear baseline classifier, in learning features for discriminating between patients with schizophrenia and healthy individuals. We tried several different architectures consisting of LSTMs as well as R-CNNs. The input data was structured as a 4-dimensional tensor (4-D brain scans: $time\times W\times H\times D$).

\textbf{Baseline Classifiers}: We compared our proposed models with the results from a linear and a RBF (radial basis function) SVM. In order to make the SVM solution feasible, we reduced the size of the input by a rate of $4\times4\times3$ voxels. This resulted in a feature vector of size 77,953 (137 time points $\times$ 569 "supervoxels") per run per subject, for a total of 380 samples (95 subjects $\times$ 4 runs).

\textbf{LSTM model}: All voxel values per brain are reshaped into a vector and fed directly into a two-layer forward LSTM model. LSTMs are known for their ability to learn long-term dependencies between inputs. In this model the spatial relationship between voxels are ignored and LSTM learns the temporal relationship between activations in different voxels. The LSTM model contains two back-to-back LSTM layers of 32 LSTM units per layer.

\textbf{Conv-LSTM model}: 3-D activations for each time frame were fed into a 3-D CNN. We used 3-dimensional convolutions followed by max-pooling to extract position and scale independent features that would generalize across individuals. Identical networks were used to process each time-frame with shared weights between them. Outputs of each CNN were then reshaped into a vector that was fed as input to the LSTM network at each time step. Figure-\ref{fig1} shows an overview of the recurrent-convolutional network used here. The CNN part learns 3-D features that are invariant to translation and scaling and reduces the dimension of the input space before feeding it into LSTM. Filters of size $3\times3\times3$ (voxels) were used in each convolutional layer. The number of filters varied according to the architecture of the model, as detailed in Table 1.
The R-CNN (2,1) contains two back-to-back convolutional layers with 16 filters in each, followed by a single convolutional layer with 32 units, and lastly two back-to-back LSTM layers with 32 units. The R-CNN (1,2) model reverses the order of the convolutional layers from R-CNN (2,1). Lastly, the R-CNN (2,2,1) model contains two back-to-back convolutional layers with 16 units per layer, followed by two back-to-back convolutional layers with 32 units each, followed by one convolutional layer with 32 units, and finished with two back-to-back LSTM layers with 32 units. 

\subsection{Training Details}
\label{training_details}

We used Adam optimizer with default parameter settings ($\beta_1=0.9, \beta_2=0.999, \epsilon=10^{-8}$) \citep{Kingma2014} and learning rate of 0.0001. We included 50 \% dropout in the input and outputs of the LSTM cells \citep{Zaremba2014}, which contained most of the tunable parameters in our models. In addition, we applied $l_2$ regularization with $\lambda=10^{-4}$ to all convolutional and fully connected weights.
3-D-convolutions of size $3\times3\times3$ with stride 1 were used. Each block of convolutions were followed by a maxpool layer of size 2 and stride 2. Size of the input to the network was $(53\times64\times37)$ voxels. 
A batch size of 64 was used for all models. Larger convolutional models were trained synchronously on 16 GPUs. In both cases, we experimented with varying number of time-frames included in each sample - viz., 16 and 64.
We used 10-fold cross-validation to evaluate the performance for each proposed model, where each fold was independently trained using different samples. Subjects in the training, evaluation, and test sets were disjoint, i.e. all samples corresponding to a subject were used only in one of the three sets. For training, samples within each batch were generated by randomly selecting time windows from different subjects. All possible time windows of size $T$ from all subjects were used for training, validation and test. We used the validation set for early stopping of training. After each training epoch, performance on the validation set was computed. Training was performed for 10 epochs and test performance was computed for the network state with highest validation score.


\subsection{Data Preparation}

Preprocessed fMRI data was normalized in three steps: For each subject/run/voxel, (i) the BOLD signal time-series was demeaned; (ii) The resulting time-series was divided by standard deviation (SD) of activation across the whole brain (all voxels and times). Next, each voxel's time-series was standardized based on the mean and SD of the voxel's activation across all subjects and runs.
Each fMRI sample (subject/run) was split into windows of 64 time points (i.e. 128 sec with a total of 137-63=74 samples), which formed the input for training, evaluation, and testing of our models. A shorter time-frame of 16 volumes (32 sec) was also tested.

\section{Results}

We investigated the effectiveness of LSTM and R-CNNs in capturing the temporal and spatial structure of the fMRI data in order to distinguish between patients and controls. The average accuracy of the baseline method was 57.89\%. 
Both LSTM and R-CNN models performed better than the baseline model (Table \ref{table_1}). Our best LSTM model performed slightly better than our R-CNN model ($ \sim 1\%$). Additionally, we explored the impact of length of the samples on the performance of the trained network . We found that larger time windows improved the accuracy in all tested models. For the R-CNN models, we experimented with the number of convolutional layers and found that the deeper CNN models did not reduce the error (and slightly increased it).

\begin{table}[h]
\centering
\caption{Comparison of different model architectures and time-windows on the AO dataset.}
\label{table_1}
\begin{tabular}{ c c c c c c c c }
\hline
\textbf{Model}           & \multicolumn{2}{c}{\textbf{LSTM}} &
\multicolumn{2}{c}{\textbf{RCNN (2, 1)}} & \multicolumn{2}{c}{\textbf{RCNN (1, 2)}} & \textbf{RCNN (2, 2, 1)} \\ \hline
\textbf{Window Size}      & 16   & 64  & 16  & 64    &16    & \ 64    & 64           \\ \hline
\textbf{Test Performance} & 60.2 \%   & 66.4\%   & 63.1\% & 64.9\% & 60.9\%  & 61.4\%  & 63.3\%  \\  \hline
\textbf{\# Conv Filters}  & - & -   & 16, 32  & 16, 32 & 16, 32 & 16, 32 & 16, 32, 32 \\ \hline
\end{tabular}
\end{table}


The best recurrent convolutional model was obtained with RCNN (2,1) containing two back-to-back convolutions in the first layer, one convolution in the second layer, and two back-to-back LSTMs in the third layer. We compared results using both 16 and 64 time-window samples while keeping the batch size fixed at 64 samples. For the model using 64 time windows, we noticed a significant improvement of 1.8\% in classification accuracy. For the LSTM model, the improvement with increasing the time window was over 6\%. Comparing the performance of the baseline models and deep learning, the test scores of LSTM and R-CNN were $\sim$ 8\% better than linear SVM (with an accuracy of 57.9\%, and false positive and negative rates of 30.6\% and 54.0\%, respectively), and ??\% better than RBF SVM (with an accuracy of 62.1\%, and false positive and negative rates of 19.9\% and 57.1\%, respectively).

\section{Discussion}

We applied several neural network architectures to learn invariant markers for schizophrenia disorder from a large-scale fMRI dataset. In particular we tried sequential LSTM networks alone and in combination with CNN input layers. While both of these methods achieved remarkable accuracy in distinguishing between patients with schizophrenia and healthy control subjects, they fell short of reaching the same level as hand-designed connectivity features (74\% in \citep{Gheiratmand2017}). This may be due to the relatively small sample size for an end-to-end deep learning framework. The 2-layer LSTM architecture, which was used in all our models, helped to learn the temporal dependencies in long time windows. The R-CNN model learned better representations compared to the LSTM model when using short time windows ($T=16$) but failed to match the LSTM performance when using the longer time windows ($T=64$). In future work, we will exploit other fMRI data subsets in the FBIRN dataset that are acquired in response to three other tasks (working memory, sensorimotor, and breath hold) to enrich the sample space. Applying transformations (such as wavelets) to fMRI data before feeding to DNN might also help improve the subtle signal in the data, compared to directly feeding the BOLD signal to the network.



\bibliography{nips_2017}
\bibliographystyle{ieeetr}

\end{document}